\documentclass[11pt,a4paper]{article}
\usepackage[margin=1in]{geometry}
\usepackage[T1]{fontenc}
\usepackage[utf8]{inputenc}
\usepackage{lmodern}
\usepackage{microtype}
\usepackage[numbers,sort&compress]{natbib}
\usepackage{graphicx}
\usepackage{amsmath,amssymb}
\usepackage{booktabs,tabularx}
\usepackage{pgfplots}
\usepackage{adjustbox}
\usepackage[hidelinks]{hyperref}
\pgfplotsset{compat=1.18}

\title{Reliability-Aware Probabilistic Fusion for Multimodal Learning with Incomplete Evidence}

\author{
    Patitapaban Palo \quad Abhisek Tiwari \quad Mattia Carletti \quad Anshul Thakur \\ \quad Xiao Gu \quad David A. Clifton \\[0.5em]
     Institute of Biomedical Engineering, \\
Department of Engineering Science, \\
University of Oxford, UK
}
\date{}
\begin{document}
\maketitle
\begin{abstract}
Multimodal inputs may be missing or degraded, raising distinct questions of evidence availability and trust. We introduce a reliability-aware Gaussian Product-of-Experts (PoE) in which an input-dependent reliability weight tempers each modality's latent precision, while missing experts are omitted. We characterize how reliability and availability affect posterior information and expert influence. Across two clinical multimodal tasks, the full framework remains competitive with complete inputs and achieves its clearest mean-AUROC advantage under modality absence, particularly on HAIM-MIMIC-MM. Ablations show a substantial gain from missing-aware training and an additional HAIM improvement from reliability weighting under matched PoE, missing-aware, and consistency training. Unseen test-time corruptions are used diagnostically and reveal modality-dependent reliability-weight and latent-uncertainty responses, including failure under severe degradation. Thus, reliability weighting provides an explicit evidence-control mechanism, while learning dependable quality signals remains an open challenge.  The implementation code is available at \url{https://github.com/patitapabanpalo/reliability-aware-probabilistic-fusion}.
\end{abstract}

\section{Introduction}
\label{sec:introduction}

Multimodal learning combines complementary information from different sources, such as images, physiological signals, structured measurements, and text, and has become increasingly important in biomedical prediction \citep{baltrusaitis2019multimodal,acosta2022multimodal}. In practice, however, multimodal inputs are rarely equally available or equally trustworthy. A modality may be entirely missing, but it may also be present yet degraded by noise, acquisition artifacts, or measurement errors. This creates two related but distinct questions: \emph{what should a model do when evidence is unavailable, and how much should it trust evidence that is available?}

Existing work has largely focused on the first question. Modality dropout trains models to tolerate absent inputs \citep{neverova2016moddrop}, while M$^3$Care uses information from similar patients to compensate for missing modalities \citep{zhang2022m3care}. More recent methods learn representations or routing mechanisms that adapt to different modality combinations. DrFuse separates shared and modality-specific information \citep{yao2024drfuse}, MUSE uses graph-based multimodal representations \citep{wu2024multimodal}, and Flex-MoE routes different availability patterns through specialized experts \citep{yun2024flexmoe}. These approaches provide effective mechanisms for incomplete inputs, but missingness is only part of the problem. An observed modality may itself contain uncertain or unreliable evidence, and representation-level fusion does not necessarily provide an explicit statistical account of how strongly such evidence should influence the prediction.

\begin{figure}[t]
    \centering
    \includegraphics[width=\linewidth]{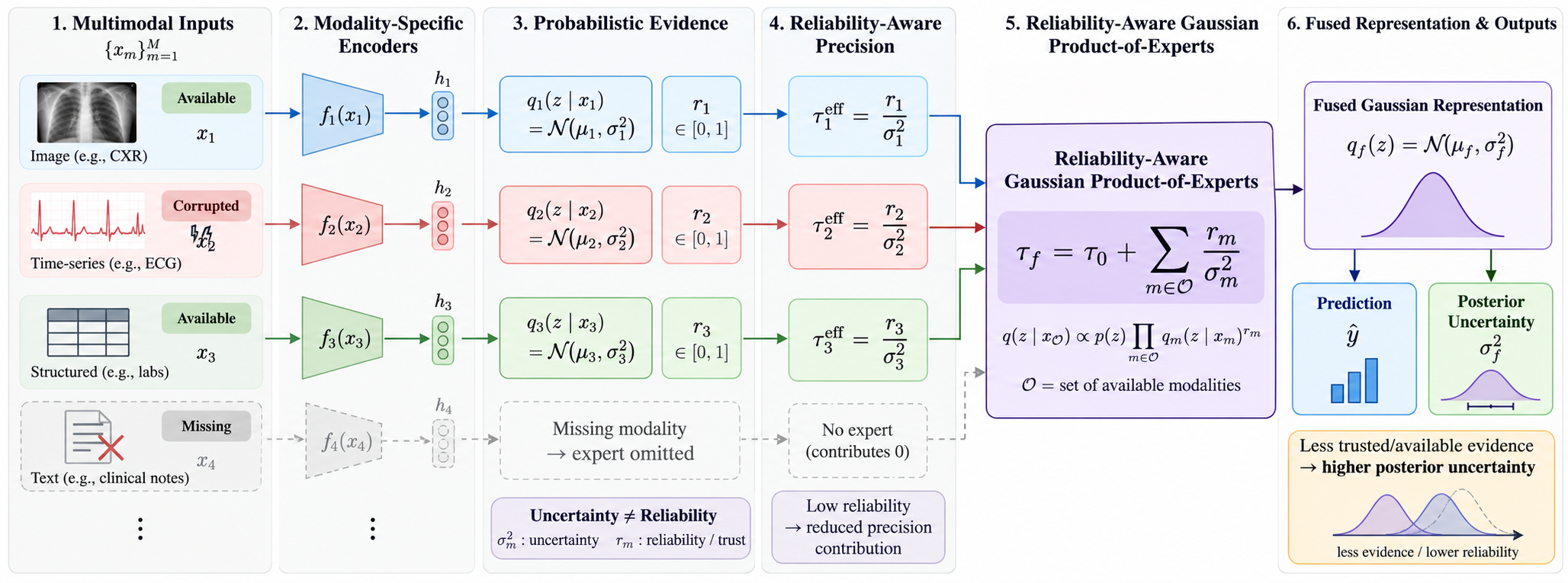}
    \caption{Framework overview. Uncertainty and reliability weights determine expert contributions; missing modalities are omitted.}
    \label{fig:framework}
\end{figure}

Probabilistic fusion provides a natural way to reason about this. Product-of-Experts (PoE) models combine evidence through the precision of individual probabilistic experts \citep{hinton2002training}, and Gaussian PoE has been used to support inference from arbitrary subsets of modalities \citep{wu2018multimodal,sutter2021generalized}. Yet precision alone need not reflect trustworthiness. An expert can assign low variance to an observation that is nevertheless corrupted or misleading, and precision miscalibration can therefore distort multimodal aggregation \citep{shi2019variational, sutter2021generalized}. This motivates separately parameterizing two operational roles:latent uncertainty describes how concentrated an expert's evidence is, whereas a reliability-interpreted weight controls how strongly that evidence enters fusion.

We introduce a reliability-aware Gaussian PoE in which an input-dependent reliability weight tempers each modality's precision while missing experts are omitted. We characterize the resulting posterior and evaluate it on cardiomegaly prediction with CXR, ECG, and laboratory measurements (Symile-MIMIC) and 48-hour mortality prediction with structured data, physiological time series, CXR, and text (HAIM-MIMIC-MM). Across five seeds, the full framework is competitive with complete inputs and shows its clearest mean-AUROC advantage under modality absence; corruption analysis shows that learned reliability-weight responses are strongly modality dependent.

Our main contributions are:
\begin{itemize}
    \item We introduce an \textbf{input-dependent reliability-weighted Gaussian PoE} that parameterizes latent variance and evidence weighting separately while supporting arbitrary observed modality subsets.
    \item We \textbf{characterize the fusion operator}, linking reliability weights and modality removal to expert influence and posterior covariance.
    \item Across two clinical multimodal tasks, we evaluate complete, missing, and corrupted inputs and \textbf{analyze both the benefits and failure modes} of learned reliability weights under degradation.
\end{itemize}

\section{Related Work}
\label{sec:related_work}

\paragraph{Learning with missing modalities.}
Multimodal models are often trained under complete observations even
though one or more modalities may be unavailable at inference time
\citep{baltrusaitis2019multimodal,wu2026missing}. Early approaches improve
robustness by randomly dropping modalities during training
\citep{neverova2016moddrop}, while other methods reconstruct or infer
missing information from the observed modalities
\citep{suzuki2016joint,vedantam2018generative,ma2021smil}.
Recent work increasingly avoids direct input reconstruction and instead
learns representations that remain useful under incomplete observations.
M$^3$Care retrieves task-relevant information from similar patients to
compensate for missing modalities \citep{zhang2022m3care}; DrFuse
separates shared and modality-specific representations
\citep{yao2024drfuse}; and MUSE uses graph-based representation learning
to accommodate missing modalities and labels \citep{wu2024multimodal}.
Mixture-of-experts approaches such as Flex-MoE further specialize
computation according to the observed modality combination
\citep{yun2024flexmoe}. These methods flexibly handle incomplete inputs through reconstruction, representation learning, or routing. Our focus is complementary: how much statistical evidence an available modality should contribute, while missing modalities contribute no expert.

\paragraph{Probabilistic fusion, uncertainty, and reliability.}
Probabilistic multimodal models provide a natural alternative to
deterministic fusion. Multimodal variational models have combined
modality-specific posteriors using products of experts, mixtures of
experts, and combinations of the two
\citep{wu2018multimodal,shi2019variational,sutter2021generalized}.
These formulations support inference from different subsets of
modalities, but the contribution of an expert is typically determined by
its learned posterior distribution. In a Gaussian Product-of-Experts,
this makes precision central to fusion, which can be problematic when
uncertainty is miscalibrated or when an observed modality is degraded.
A related line of work studies trustworthy multi-view learning through
evidential uncertainty. Trusted Multi-View Classification uses
Dirichlet evidence and Dempster--Shafer fusion to quantify uncertainty
and improve robustness to unreliable views
\citep{han2021trusted,han2023dynamic}, while subsequent evidential fusion
methods similarly model uncertainty when combining heterogeneous sources
\citep{zhang2024dual}. These approaches motivate trust-aware fusion but typically express trust through predictive evidence or uncertainty itself. We instead parameterize two roles: latent variance describes concentration of an expert's evidence, whereas an input-dependent reliability weight tempers how strongly that evidence enters the PoE. This is best viewed as a reliability-interpreted, input-dependent tempering of Gaussian experts rather than new PoE algebra; its value is the separate parameterization of these roles and their use under variable modality availability.

\section{Reliability-Aware Probabilistic Fusion}
\label{sec:method}

Figure~\ref{fig:framework} treats each available modality as probabilistic evidence with two separately parameterized roles: latent uncertainty controls evidence concentration, while reliability weighting controls its contribution to fusion. The formulation supports arbitrary modality types and subsets.

\subsection{Problem Formulation}

Let $\mathcal{X}=\{x_1,\ldots,x_M\}$ denote the $M$ possible modalities
for an example, and let
$\mathcal{O}\subseteq\{1,\ldots,M\}$ denote the subset observed at
inference time. Given the available inputs
$x_{\mathcal{O}}=\{x_m:m\in\mathcal{O}\}$, our goal is to predict a
target $y$ while representing the uncertainty associated with the
available evidence.

Each observed modality is first mapped to a common representation,
\begin{equation}
    h_m=f_m(x_m;\theta_m),
\end{equation}
where $f_m$ is modality-specific. This allows the feature extractor to
reflect the structure of each input while keeping the fusion mechanism
shared across modalities. Similar modality-specific encoding followed by
a shared latent representation is common in multimodal latent-variable
models \citep{suzuki2016joint,wu2018multimodal,shi2019variational}.

\subsection{Probabilistic Modality Evidence}

Rather than representing modality $m$ by a single deterministic vector,
we model its latent evidence as a diagonal Gaussian expert,
\begin{equation}
    q_m(\mathbf{z}\mid x_m)
    =
    \mathcal{N}\!\left(
        \mathbf{z};
        \boldsymbol{\mu}_m,
        \operatorname{diag}(\boldsymbol{\sigma}_m^2)
    \right),
\end{equation}
where $\boldsymbol{\mu}_m\in\mathbb{R}^d$ and
$\log\boldsymbol{\sigma}_m^2\in\mathbb{R}^d$ are predicted from $h_m$.
We denote the corresponding element-wise precision vector by
$\boldsymbol{\tau}_m=\boldsymbol{\sigma}_m^{-2}$. Gaussian
modality-specific posteriors provide a convenient way to combine
different subsets of observations and have been widely used in
multimodal variational inference
\citep{wu2018multimodal,shi2019variational,sutter2021generalized}.

The variance $\sigma_m^2$ describes uncertainty in the latent evidence.
We additionally predict a scalar reliability-interpreted evidence weight
$r_m$,
\begin{equation}
    r_m =
    \operatorname{sigmoid}
    \left(
    g_{r,m}
    ([\boldsymbol{\mu}_m;
    \log\boldsymbol{\sigma}_m^2])
    \right),
    \qquad r_m\in(0,1),
    \label{eq:reliability}
\end{equation}
where $g_{r,m}$ is a modality-specific reliability head. The scalar is
parameterized separately from latent variance and controls how strongly
modality $m$ enters fusion. Because only their product determines effective
precision, task supervision alone does not identify $r_m$ as a calibrated
quality probability.

\subsection{Reliability-Aware Product-of-Experts}

Product-of-Experts (PoE) fusion provides a natural mechanism for
combining probabilistic evidence \citep{hinton2002training,wu2018multimodal}.
For a standard Gaussian PoE, an expert with smaller variance contributes
greater precision to the fused posterior. We extend this mechanism by
tempering each expert according to its learned reliability weight:
\begin{equation}
    q(z\mid x_{\mathcal O})
    \propto
    p(z)
    \prod_{m\in\mathcal O}
    q_m(z\mid x_m)^{r_m},
    \label{eq:reliability_poe}
\end{equation}
where $p(\mathbf{z})=\mathcal{N}(\mathbf{0},I)$ is a Gaussian prior.
Each powered expert $q_m(\mathbf{z}\mid x_m)^{r_m}$ is understood up to
normalization; raising a Gaussian expert to a nonnegative power $r_m$
scales its precision by $r_m$. For diagonal Gaussian
experts, the effective precision contributed by modality $m$ is
\begin{equation}
    \tilde{\boldsymbol{\tau}}_m
    =
    r_m\boldsymbol{\tau}_m
    =
    r_m\boldsymbol{\sigma}_m^{-2},
    \label{eq:effective_precision}
\end{equation}
The fused precision and mean therefore have the closed forms
\begin{align}
    \boldsymbol{\tau}_f
    &=
    \boldsymbol{\tau}_0+
    \sum_{m\in\mathcal O}
    r_m\boldsymbol{\tau}_m,
    \label{eq:fused_precision}\\
    \boldsymbol{\mu}_f
    &=
    \boldsymbol{\tau}_f^{-1}
    \odot
    \left(
        \boldsymbol{\tau}_0\odot\boldsymbol{\mu}_0
        +
        \sum_{m\in\mathcal O}
        r_m\boldsymbol{\tau}_m\odot\boldsymbol{\mu}_m
    \right),
    \label{eq:fused_mean}
\end{align}

where $\odot$ denotes element-wise multiplication and the inverse of
$\boldsymbol{\tau}_f$ is also taken element-wise.
With a standard normal prior,
$\boldsymbol{\mu}_0=\mathbf{0}$ and
$\boldsymbol{\tau}_0=\mathbf{1}$. This formulation
preserves Gaussian PoE aggregation while separately parameterizing
expert precision and an input-dependent tempering weight.
Setting $r_m=1$ recovers the standard Gaussian PoE, whereas the limiting
case $r_m\rightarrow0$ removes the expert's precision contribution. Although
our sigmoid parameterization yields $r_m\in(0,1)$, the fusion operator is
well defined for $r_m\geq0$. We formalize these properties in
Section~\ref{sec:theory}.

\subsection{Learning Reliability}

The reliability head is learned end-to-end through supervised fusion, without corruption-ranking supervision. Thus, $r_m$ is an input-dependent evidence weight, not a calibrated quality probability. Since fusion depends only on $r_m\boldsymbol{\sigma}_m^{-2}$, reliability and latent variance are not identifiable from task supervision alone; Section~\ref{sec:reliability_analysis} examines their behavior under held-out degradation.

\subsection{Learning with Missing Modalities}

The PoE formulation does not require imputation when a modality is
missing: its expert is simply absent from Eq.~\eqref{eq:reliability_poe}.
During training, we expose the model to different observed subsets
$\mathcal{S}\subseteq\{1,\ldots,M\}$, similar in spirit to modality
dropout and subset-based multimodal learning
\citep{wu2018multimodal,sutter2021generalized}.
The corresponding posterior is formed only from the available experts.

We optimize predictions from both complete and sampled incomplete
modality sets. Inspired by teacher--student consistency and self-distillation objectives
\citep{hinton2015distilling,tarvainen2017mean}, we use the
complete-modality prediction as a stop-gradient target for predictions
from incomplete subsets:
\begin{equation}
    \mathcal{L}_{\mathrm{cons}}
    =
    \left\|
        \hat{y}_{\mathcal S}
        -
        \operatorname{sg}(\hat{y}_{\mathrm{full}})
    \right\|_2^2,
    \label{eq:consistency}
\end{equation}
where $\operatorname{sg}(\cdot)$ denotes stop-gradient. The complete
prediction therefore acts as a training target without being moved by
the incomplete branch.

The final objective is
\begin{equation}
    \mathcal{L}
    =
    \mathcal{L}_{\mathrm{task}}
    +\lambda_{\mathrm{uni}}\mathcal{L}_{\mathrm{uni}}
    +\lambda_{\mathrm{cons}}\mathcal{L}_{\mathrm{cons}},
    \label{eq:full_objective}
\end{equation}
Here, $\mathcal{L}_{\mathrm{task}}$ denotes the supervised prediction
loss over the complete and sampled incomplete modality sets, while
$\mathcal{L}_{\mathrm{uni}}$ applies the supervised loss to each
individual modality to preserve unimodal predictive information.
For our binary prediction tasks, the supervised objective is a
class-weighted binary cross-entropy loss. This gives the general training objective; dataset-specific experiments
may omit auxiliary terms when they are not used. Dataset-specific architectures, subset-sampling
details, loss weights, and optimization settings are provided in
Section~\ref{app:implementation}.

\section{Statistical Properties}
\label{sec:theory}

We characterize operator-level effects of reliability and availability; these results do not assume that learned weights detect corruption. Proofs are in Section~\ref{app:proofs}; the fusion and information-ordering results extend from our diagonal implementation to positive-definite Gaussian covariances.

Let the prior and modality experts be Gaussian,
$p(\mathbf{z})=\mathcal{N}(\boldsymbol{\mu}_0,\Sigma_0)$ and
$q_m(\mathbf{z}\mid x_m) = \mathcal{N}(\boldsymbol{\mu}_m,\Sigma_m)$, with positive-definite
precision matrices $\Lambda_0=\Sigma_0^{-1}$ and
$\Lambda_m=\Sigma_m^{-1}$, and reliability $r_m\geq0$.

\paragraph{Proposition 1 (Reliability-weighted Gaussian fusion).}
For an observed modality set $\mathcal{O}$, the reliability-weighted
Product-of-Experts in Eq.~\eqref{eq:reliability_poe} is Gaussian with
\begin{align}
    \Lambda_{\mathcal O}
    &=
    \Lambda_0+
    \sum_{m\in\mathcal O} r_m\Lambda_m,
    \label{eq:theory_precision}\\
    \boldsymbol{\mu}_{\mathcal O}
    &=
    \Lambda_{\mathcal O}^{-1}
    \left(
        \Lambda_0\boldsymbol{\mu}_0+
        \sum_{m\in\mathcal O}
        r_m\Lambda_m\boldsymbol{\mu}_m
    \right).
    \label{eq:theory_mean}
\end{align}
Thus, reliability scales the information contributed by an expert rather
than its latent estimate directly. Standard Gaussian PoE is recovered
when $r_m=1$ for all $m$, while $r_m=0$ gives zero information
contribution from expert $m$.

\paragraph{Proposition 2 (Reliability--uncertainty monotonicity).}
Consider one latent coordinate with fused variance
\begin{equation}
    V_j =
    \left(
        \tau_{0j}+
        \sum_{m\in\mathcal O}r_m\tau_{mj}
    \right)^{-1}.
\end{equation}
For any expert $k$,
\begin{equation}
    \frac{\partial V_j}{\partial r_k}
    =
    -\frac{\tau_{kj}}
    {\left(
        \tau_{0j}+\sum_m r_m\tau_{mj}
    \right)^2}
    < 0.
    \label{eq:reliability_variance_derivative}
\end{equation}
Since $\tau_{kj}>0$, $V_j$ is strictly decreasing in $r_k$.
Thus, reducing the reliability weight assigned to an available expert increases
posterior variance for that coordinate. In other words, decreasing an available
source's reliability weight cannot make the fused representation more certain.

\paragraph{Proposition 3 (Information monotonicity under missingness).}
Let $\mathcal{S}\subseteq\mathcal{O}$ be two observed modality sets,
with the reliability scores of retained experts fixed. Then
\begin{equation}
    \Lambda_{\mathcal S}
    \preceq
    \Lambda_{\mathcal O},
    \qquad
    \Sigma_{\mathcal S}
    \succeq
    \Sigma_{\mathcal O},
    \label{eq:missing_information_order}
\end{equation}
where $\preceq$ denotes the Loewner partial order. Therefore, removing
nonnegative-precision evidence cannot reduce posterior covariance. This
provides a direct statistical interpretation of missing-modality
inference: as fewer experts are available, the fused posterior cannot
become more informative solely through their removal.

\paragraph{Proposition 4 (Conditional expert influence).}
For diagonal Gaussian experts, holding $r_k$ and $\tau_{kj}$ fixed,
the partial sensitivity of fused coordinate $\mu_{f,j}$ to expert mean
$\mu_{k,j}$ is
\begin{equation}
    \left.
    \frac{\partial\mu_{f,j}}
         {\partial\mu_{k,j}}
    \right|_{r_k,\tau_{kj}}
    =
    \frac{r_k\tau_{kj}}
    {\tau_{0j}+\sum_m r_m\tau_{mj}}
    \equiv \alpha_{kj}.
    \label{eq:expert_influence}
\end{equation}
The coefficient $\alpha_{kj}\in[0,1)$ is the expert's normalized effective
precision. Consequently, as $r_k\rightarrow0$,
$\alpha_{kj}\rightarrow0$: perturbations to an expert's latent estimate
have vanishing influence on the fused mean.

Together, Propositions 2--4 show that decreasing $r_m$ increases posterior uncertainty and limits expert influence; as $r_m\rightarrow0$, its direct information contribution vanishes. Missingness remains distinct from unreliability: a missing expert is absent, whereas an unreliable one is observed but down-weighted. Whether degradation actually reduces learned $r_m$ is evaluated empirically.

\subsection{Detailed Proofs of Statistical Properties}
\label{app:proofs}

This section provides complete proofs of the results stated in
Section~\ref{sec:theory}. We use the notation introduced in the main
text. Let
\[
p(\mathbf z)
=
\mathcal N(\mathbf z;\boldsymbol{\mu}_0,\Sigma_0)
\]
denote the Gaussian prior and
\[
q_m(\mathbf z\mid x_m)
=
\mathcal N(\mathbf z;\boldsymbol{\mu}_m,\Sigma_m)
\]
the Gaussian expert associated with modality $m$. We write
$\Lambda_0=\Sigma_0^{-1}$ and
$\Lambda_m=\Sigma_m^{-1}$ for their precision matrices, with
$r_m\geq0$ denoting modality reliability.

\subsubsection{Proof of Proposition 1}
\label{app:proof_prop1}

For an observed modality set $\mathcal O$, the reliability-weighted
Product-of-Experts is
\begin{equation}
q(\mathbf z\mid x_{\mathcal O})
\propto
p(\mathbf z)
\prod_{m\in\mathcal O}
q_m(\mathbf z\mid x_m)^{r_m}.
\label{eq:app_poe}
\end{equation}
The Gaussian prior can be written in information form as
\begin{equation}
p(\mathbf z)
\propto
\exp\left\{
-\frac{1}{2}
(\mathbf z-\boldsymbol{\mu}_0)^\top
\Lambda_0
(\mathbf z-\boldsymbol{\mu}_0)
\right\}.
\end{equation}
Similarly, raising expert $m$ to the nonnegative power $r_m$ gives
\begin{align}
q_m(\mathbf z\mid x_m)^{r_m}
&\propto
\exp\left\{
-\frac{r_m}{2}
(\mathbf z-\boldsymbol{\mu}_m)^\top
\Lambda_m
(\mathbf z-\boldsymbol{\mu}_m)
\right\}.
\label{eq:app_powered_expert}
\end{align}
Thus, exponentiation by $r_m$ scales the expert precision from
$\Lambda_m$ to $r_m\Lambda_m$.

Taking the logarithm of Eq.~\eqref{eq:app_poe} and discarding terms that
do not depend on $\mathbf z$ yields
\begin{align}
\log q(\mathbf z\mid x_{\mathcal O})
&=
-\frac{1}{2}
(\mathbf z-\boldsymbol{\mu}_0)^\top
\Lambda_0
(\mathbf z-\boldsymbol{\mu}_0)
\nonumber\\
&\quad
-\frac{1}{2}
\sum_{m\in\mathcal O}
r_m
(\mathbf z-\boldsymbol{\mu}_m)^\top
\Lambda_m
(\mathbf z-\boldsymbol{\mu}_m)
+C,
\label{eq:app_log_poe}
\end{align}
where $C$ collects terms independent of $\mathbf z$.

Expanding the quadratic terms and collecting coefficients of
$\mathbf z$ gives
\begin{align}
\log q(\mathbf z\mid x_{\mathcal O})
&=
-\frac{1}{2}
\mathbf z^\top
\left(
\Lambda_0+
\sum_{m\in\mathcal O}r_m\Lambda_m
\right)
\mathbf z
\nonumber\\
&\quad+
\mathbf z^\top
\left(
\Lambda_0\boldsymbol{\mu}_0+
\sum_{m\in\mathcal O}
r_m\Lambda_m\boldsymbol{\mu}_m
\right)
+C'.
\label{eq:app_collect_terms}
\end{align}

Define
\begin{equation}
\Lambda_{\mathcal O}
=
\Lambda_0+
\sum_{m\in\mathcal O}r_m\Lambda_m.
\label{eq:app_fused_precision}
\end{equation}
Because $\Lambda_0\succ0$, $\Lambda_m\succ0$, and $r_m\geq0$,
we have $\Lambda_{\mathcal O}\succ0$. Completing the square therefore
shows that Eq.~\eqref{eq:app_collect_terms} is the log-density of a
Gaussian distribution with covariance
\begin{equation}
\Sigma_{\mathcal O}
=
\Lambda_{\mathcal O}^{-1}
\end{equation}
and mean
\begin{equation}
\boldsymbol{\mu}_{\mathcal O}
=
\Lambda_{\mathcal O}^{-1}
\left(
\Lambda_0\boldsymbol{\mu}_0+
\sum_{m\in\mathcal O}
r_m\Lambda_m\boldsymbol{\mu}_m
\right).
\label{eq:app_fused_mean}
\end{equation}
This proves Proposition~1.

Two useful special cases follow immediately. If $r_m=1$ for every
available modality, Eq.~\eqref{eq:app_fused_precision} reduces to the
standard Gaussian Product-of-Experts precision. If $r_k=0$, expert $k$
contributes zero precision and zero precision-weighted mean to the
fused posterior.

\subsubsection{Proof of Proposition 2}
\label{app:proof_prop2}

For the diagonal Gaussian model used in our implementation, consider
latent coordinate $j$. Let $\tau_{0j}>0$ denote the prior precision and
$\tau_{mj}>0$ the precision contributed by modality $m$ before
reliability weighting. From Proposition~1, the fused precision is
\begin{equation}
\tau_{f,j}
=
\tau_{0j}
+
\sum_{m\in\mathcal O}
r_m\tau_{mj}.
\end{equation}
Hence the corresponding posterior variance is
\begin{equation}
V_j
=
\frac{1}
{\tau_{0j}+\sum_{m\in\mathcal O}r_m\tau_{mj}}.
\label{eq:app_variance}
\end{equation}

Differentiating Eq.~\eqref{eq:app_variance} with respect to the
reliability of expert $k$ gives
\begin{align}
\frac{\partial V_j}{\partial r_k}
&=
-\left(
\tau_{0j}+\sum_{m\in\mathcal O}r_m\tau_{mj}
\right)^{-2}
\tau_{kj}
\\
&=
-\frac{\tau_{kj}}
{\left(
\tau_{0j}+\sum_{m\in\mathcal O}r_m\tau_{mj}
\right)^2}.
\end{align}
Since $\tau_{kj}>0$ and the denominator is strictly positive,
\begin{equation}
\frac{\partial V_j}{\partial r_k}<0.
\end{equation}
Therefore $V_j$ is strictly decreasing in $r_k$. Equivalently,
reducing the reliability assigned to an available expert increases
posterior variance for that latent coordinate. This proves
Proposition~2.

\subsubsection{Proof of Proposition 3}
\label{app:proof_prop3}

Let $\mathcal S\subseteq\mathcal O$ be two observed modality sets, and
hold the reliability scores of all retained experts fixed. From
Proposition~1,
\begin{align}
\Lambda_{\mathcal O}
&=
\Lambda_0+
\sum_{m\in\mathcal O}r_m\Lambda_m,\\
\Lambda_{\mathcal S}
&=
\Lambda_0+
\sum_{m\in\mathcal S}r_m\Lambda_m.
\end{align}
Their difference is
\begin{equation}
\Lambda_{\mathcal O}-\Lambda_{\mathcal S}
=
\sum_{m\in\mathcal O\setminus\mathcal S}
r_m\Lambda_m.
\label{eq:app_precision_difference}
\end{equation}
For every $m$, $r_m\geq0$ and $\Lambda_m\succ0$. Consequently,
$r_m\Lambda_m\succeq0$, and therefore
\begin{equation}
\Lambda_{\mathcal O}-\Lambda_{\mathcal S}\succeq0.
\end{equation}
It follows that
\begin{equation}
\Lambda_{\mathcal S}
\preceq
\Lambda_{\mathcal O}.
\label{eq:app_precision_order}
\end{equation}

Both precision matrices are positive definite because they contain the
positive-definite prior precision $\Lambda_0$. Matrix inversion reverses
the Loewner order for positive-definite matrices. Applying this property
to Eq.~\eqref{eq:app_precision_order} gives
\begin{equation}
\Lambda_{\mathcal S}^{-1}
\succeq
\Lambda_{\mathcal O}^{-1}.
\end{equation}
Since
$\Sigma_{\mathcal S}=\Lambda_{\mathcal S}^{-1}$ and
$\Sigma_{\mathcal O}=\Lambda_{\mathcal O}^{-1}$,
\begin{equation}
\Sigma_{\mathcal S}
\succeq
\Sigma_{\mathcal O}.
\end{equation}
Thus, removing nonnegative-precision experts cannot decrease posterior
covariance. This proves Proposition~3.

\subsubsection{Proof of Proposition 4}
\label{app:proof_prop4}

For the diagonal model, consider latent coordinate $j$. From
Proposition~1, the fused mean can be written as
\begin{equation}
\mu_{f,j}
=
\frac{
\tau_{0j}\mu_{0j}
+
\sum_m r_m\tau_{mj}\mu_{m,j}
}{
\tau_{0j}
+
\sum_m r_m\tau_{mj}
}.
\label{eq:app_scalar_mean}
\end{equation}
We consider the partial sensitivity to $\mu_{k,j}$ while holding
$r_k$ and $\tau_{kj}$ fixed. Under this conditioning, the denominator
of Eq.~\eqref{eq:app_scalar_mean} does not depend on $\mu_{k,j}$.
Differentiating therefore gives
\begin{equation}
\left.
\frac{\partial\mu_{f,j}}
{\partial\mu_{k,j}}
\right|_{r_k,\tau_{kj}}
=
\frac{
r_k\tau_{kj}
}{
\tau_{0j}+\sum_m r_m\tau_{mj}
}
\equiv
\alpha_{kj}.
\label{eq:app_influence}
\end{equation}

All terms in the denominator are nonnegative and the prior precision is
strictly positive. Hence
\begin{equation}
0\leq\alpha_{kj}<1.
\end{equation}
In particular,
\begin{equation}
\lim_{r_k\rightarrow0}\alpha_{kj}=0.
\end{equation}
Thus, conditional on fixed expert precision and reliability, the
sensitivity of the fused mean to expert $k$ vanishes as its reliability
approaches zero. This proves Proposition~4.

\paragraph{Remark.}
The conditioning in Proposition~4 is important because the learned
reliability $r_k$ is itself predicted from modality-specific evidence
in our model. Proposition~4 characterizes the direct influence induced
by the fusion rule; it does not include indirect changes that may occur
if perturbing $\boldsymbol{\mu}_k$ also changes the output of the
reliability network.

\section{Experimental Setup}
\label{sec:experiments}

We evaluate the proposed framework on two multimodal clinical prediction
settings that differ in both task and modality composition. Our
experiments address three questions: (i) whether probabilistic fusion
remains competitive when all modalities are available, (ii) how models
behave when one or more modalities are missing, and (iii) how robust
fusion remains when an observed modality is corrupted.

\subsection{Datasets and Tasks}

\paragraph{Symile-MIMIC.}
We study cardiomegaly classification from CXR, ECG, and laboratory measurements using the released partitions after retaining examples with valid labels. Cohort construction, split sizes, and preprocessing are provided in Section~\ref{app:symile_data}.

\paragraph{HAIM-MIMIC-MM.}
We study 48-hour mortality classification from structured data, physiological time series, CXR, and clinical text using patient-level splits; observation-level prevalence is 2.73\%. Experiments use distributed precomputed representations associated with CXR-anchored observations. Cohort construction and preprocessing are detailed in Section~\ref{app:haim_data}.

\subsection{Comparison Methods}

We compare with unimodal models, early/late fusion, mean latent fusion, Gaussian and reliability-weighted PoE, modality dropout \citep{neverova2016moddrop}, and recent incomplete-modality methods DrFuse \citep{yao2024drfuse}, M$^3$Care \citep{zhang2022m3care}, MUSE \citep{wu2024multimodal}, and Flex-MoE \citep{yun2024flexmoe}. Because the published methods assume different modality sets and input representations, we instantiate their defining mechanisms with dataset-specific encoders and interfaces while following their standard methodological training procedures. Details are provided in Section~\ref{app:baselines}.

\subsection{Missing and Corrupted Modalities}

We evaluate every non-empty incomplete modality subset (6 for Symile; 14 for HAIM); the proposed model forms its posterior only from available experts. For corruption, one modality is degraded at a time at three severities. No reported model is trained on the synthetic corruption operators used for evaluation. Operators, matched HAIM perturbations, and exact parameters are given in Section~\ref{app:corruptions}. Each comparison method follows its standard missing-modality mechanism and training procedure rather than being forced into a common objective; Section~\ref{sec:ablation} separately isolates components of our framework.

\subsection{Evaluation Protocol}

We report AUROC and AUPRC in the main analysis; Brier scores are reported in the detailed results as supplementary probability-error measures and are not interpreted as calibration because training uses class-weighted BCE. Final comparisons use five independent seeds. For aggregate missingness and corruption results, conditions are averaged within each seed before computing mean $\pm$ sample SD across seeds. A paired seed-level uncertainty analysis for the principal HAIM missing-modality comparison is reported in Section~\ref{app:haim_paired_missing}. Further training and model-selection details are in Section~\ref{app:implementation}.

\subsection{Implementation and Training Details}
\label{app:implementation}

We provide implementation details for the proposed model on
Symile-MIMIC and HAIM-MIMIC-MM. Both datasets use the same
probabilistic fusion mechanism, while modality encoders and training
settings are adapted to the corresponding inputs and tasks.

\subsubsection{Model Architectures}
\label{app:model_architectures}

Both implementations map modality-specific inputs to a
128-dimensional embedding and use a 32-dimensional Gaussian latent
space. For each modality, separate linear projections predict
$\boldsymbol{\mu}_m$ and $\log\boldsymbol{\sigma}_m^2$, with
log-variances clipped to $[-6,4]$ for numerical stability.

The reliability head takes
$[\boldsymbol{\mu}_m;\log\boldsymbol{\sigma}_m^2]$ as input and uses
a $64\!\rightarrow\!32\!\rightarrow\!1$ MLP with ReLU and sigmoid
output. The prediction head maps the fused latent mean through a
$32\!\rightarrow\!32\!\rightarrow\!1$ MLP with ReLU activation.

Dataset-specific modality encoders are summarized in
Table~\ref{tab:encoders}.

\begin{table}[t]
\centering
\caption{Dataset-specific modality encoders.}
\label{tab:encoders}
\small
\setlength{\tabcolsep}{3pt}
\begin{tabularx}{\linewidth}{lX}
\toprule
\textbf{Input} & \textbf{Encoder}\\
\midrule
Symile CXR &
Conv2D channels $32,64,128,128$; adaptive pooling\\
Symile ECG &
Conv1D channels $32,64,128$; kernels $15,9,7$\\
Symile Labs &
$100\!\rightarrow\!256\!\rightarrow\!128$ MLP\\
HAIM modalities &
$d_m\!\rightarrow\!256\!\rightarrow\!128$ MLP; LN, ReLU,
dropout $0.20$\\
\bottomrule
\end{tabularx}
\end{table}

For Symile, the CXR encoder uses a $7\times7$ stride-2 first
convolution followed by $3\times3$ convolutional blocks, batch
normalization, ReLU, and adaptive average pooling. The ECG encoder
operates on the 12-channel waveform using one-dimensional convolutions,
batch normalization, ReLU, temporal pooling, and adaptive average
pooling. Laboratory values are concatenated with 50 corresponding
missingness indicators, producing a 100-dimensional input.

For HAIM, $d_m$ corresponds to the input dimension of each
representation-level modality: 1,542 structured, 451 time-series,
2,084 CXR, and 2,304 text features. Each modality uses the same MLP
structure shown in Table~\ref{tab:encoders}, with layer normalization
after each linear projection.

\subsubsection{Training Objectives}
\label{app:training_objectives}

Symile uses the task, unimodal, and consistency terms in Eq.~\eqref{eq:full_objective}. Reliability is learned end-to-end through its contribution to fusion; the final reported model does not use corruption-ranking supervision. Its task loss equally weights predictions from the complete and sampled incomplete modality sets:
\begin{equation}
    \mathcal{L}_{\mathrm{task}}
    =
    \frac{1}{2}\mathcal{L}_{\mathrm{BCE}}^{\mathrm{full}}
    +
    \frac{1}{2}\mathcal{L}_{\mathrm{BCE}}^{\mathrm{subset}}.
\end{equation}
The unimodal auxiliary loss applies the supervised objective separately
to CXR, ECG, and laboratory representations. The consistency loss compares
the sampled-subset prediction with the detached complete-modality
prediction.

HAIM Full-F uses reliability-weighted PoE, missing-modality-aware
mortality training, and complete-to-incomplete consistency, without the
unimodal auxiliary term. As on Symile, reliability is learned through its effect on
probabilistic fusion and the supervised task objective:
\begin{equation}
    \mathcal{L}_{\mathrm{HAIM}}
    =
    \mathcal{L}_{\mathrm{mort}}
    +
    \lambda_{\mathrm{cons}}\mathcal{L}_{\mathrm{cons}}.
\end{equation}
Both datasets use class-weighted binary cross-entropy for their
supervised prediction objectives.

\subsubsection{Missing-Modality Sampling}
\label{app:subset_sampling}

For Symile, training samples from all seven non-empty modality
configurations. The complete-modality pattern is sampled with
probability $0.25$, each of the three single-missing patterns with
probability $0.15$, and each of the three two-missing patterns with
probability $0.10$.

HAIM contains 15 non-empty modality configurations. For the supervised
mortality branch, the number of missing modalities is sampled as
\begin{equation}
\begin{aligned}
P(0\text{ missing}) &= 0.50, &
P(1\text{ missing}) &= 0.20,\\
P(2\text{ missing}) &= 0.20, &
P(3\text{ missing}) &= 0.10.
\end{aligned}
\end{equation}
Conditional on this number, the missing modalities are sampled
uniformly without replacement.

HAIM uses a separate incomplete subset for the consistency branch. The
number of missing modalities is sampled uniformly from $\{1,2,3\}$,
after which the corresponding modalities are sampled uniformly. Thus,
the consistency objective is never applied to the complete-modality
case.

\subsubsection{Optimization and Model Selection}
\label{app:optimization}

The principal architecture and training hyperparameters are summarized
in Table~\ref{tab:training_hyperparameters}.

\begin{table}[t]
\centering
\caption{Training hyperparameters for the proposed models.}
\label{tab:training_hyperparameters}
\small
\setlength{\tabcolsep}{3pt}
\begin{adjustbox}{max width=\linewidth}
\begin{tabular}{lcc}
\toprule
\textbf{Setting} & \textbf{Symile} & \textbf{HAIM}\\
\midrule
Embedding dimension & 128 & 128\\
Latent dimension & 32 & 32\\
Batch size & 16 & 256\\
Optimizer & Adam & AdamW\\
Learning rate & $10^{-3}$ & $10^{-3}$\\
Weight decay & 0 & $10^{-4}$\\
Max. epochs & 10 & 50\\
Early stopping & -- & 8 epochs\\
Gradient clip & 5.0 & 5.0\\
$\lambda_{\mathrm{uni}}$ & 0.50 & --\\
$\lambda_{\mathrm{cons}}$ & 0.25 & 0.10\\
Log-var. range & $[-6,4]$ & $[-6,4]$\\
\bottomrule
\end{tabular}
\end{adjustbox}
\end{table}

Symile is trained for 10 epochs. HAIM uses early stopping with patience
8 based on validation AUROC and retains the checkpoint with the highest
validation AUROC. Gradients are clipped to maximum norm 5 in both
settings.

Final results are aggregated over five independently trained seeds.
Python, NumPy, and PyTorch random-number generators are seeded for each
run. Dataset construction and preprocessing are described in
Section~\ref{app:data}, and the exact corruption protocols are given
in Section~\ref{app:corruptions}.

\subsection{Corruption Protocols}
\label{app:corruptions}

We evaluate robustness using controlled corruption at three severity
levels. One modality is corrupted at a time while all others remain
unchanged. We distinguish corruption from missingness: a missing
modality contributes no expert, whereas a corrupted modality remains
observed and may contribute unreliable evidence.

\subsubsection{Symile-MIMIC Corruptions}
\label{app:symile_corruptions}

Because Symile contains heterogeneous inputs, we use modality-specific
corruptions for CXR, ECG, and laboratory measurements. Severity
$s\in\{1,2,3\}$ denotes mild, moderate, and severe corruption,
respectively; $s=0$ denotes the unmodified input.

\paragraph{CXR.}
CXR corruption combines spatial smoothing and additive Gaussian noise:
\begin{equation}
    \widetilde{x}^{(s)}
    =
    (1-\beta_s)x
    +
    \beta_s\mathcal{B}_{k_s}(x)
    +
    \eta_s\epsilon,
    \qquad
    \epsilon\sim\mathcal{N}(0,I),
    \label{eq:cxr_corruption}
\end{equation}
where $\mathcal{B}_{k_s}$ denotes average pooling with kernel size
$k_s$, unit stride, and padding that preserves the image dimensions.
The severity parameters are given in
Table~\ref{tab:symile_corruption}.

\begin{table}[t]
\centering
\caption{Symile-MIMIC corruption parameters.}
\label{tab:symile_corruption}
\small
\setlength{\tabcolsep}{4pt}
\begin{adjustbox}{max width=\linewidth}
\begin{tabular}{lccc}
\toprule
\textbf{Severity} & \textbf{CXR $(k_s,\beta_s,\eta_s)$}
& \textbf{ECG $\eta_s$} & \textbf{Labs $\alpha_s$}\\
\midrule
Mild     & $(3,0.25,0.02)$  & 0.05 & 0.25\\
Moderate & $(7,0.55,0.05)$  & 0.15 & 0.60\\
Severe   & $(13,0.85,0.10)$ & 0.30 & 1.00\\
\bottomrule
\end{tabular}
\end{adjustbox}
\end{table}

\paragraph{ECG.}
ECG corruption adds Gaussian noise followed by clipping to the
normalized input range:
\begin{equation}
    \widetilde{x}^{(s)}
    =
    \operatorname{clip}
    (x+\eta_s\epsilon,-1,1),
    \qquad
    \epsilon\sim\mathcal{N}(0,I),
    \label{eq:ecg_corruption}
\end{equation}
with $\eta_s\in\{0.05,0.15,0.30\}$ for mild, moderate, and severe
corruption.

\paragraph{Laboratory measurements.}
Observed normalized laboratory values are progressively replaced by
independent random values:
\begin{equation}
    \widetilde{x}^{(s)}
    =
    (1-\alpha_s)x+\alpha_su,
    \qquad
    u\sim\operatorname{Uniform}(0,1),
    \label{eq:lab_corruption}
\end{equation}
where $\alpha_s\in\{0.25,0.60,1.00\}$. At severe corruption, the
laboratory-value vector is fully replaced by random values. Laboratory
missingness indicators are retained unchanged.

These corruption operators are used only for held-out robustness evaluation of the final reported models; they are not used as corruption augmentation or reliability supervision during training. The CXR protocol was also checked during development to verify that input-space distortion increased with severity.

\subsubsection{HAIM-MIMIC-MM Corruptions}
\label{app:haim_corruptions}

HAIM uses precomputed representations, so the same feature-space
corruption is applied to structured, time-series, CXR, and text
representations. For representation vector $x$,
\begin{equation}
\begin{aligned}
    \widetilde{x}^{(s)}
    &= x\odot b_s+\eta_s\epsilon,\\
    b_{s,j} &\sim \operatorname{Bernoulli}(1-p_s),
    \qquad
    \epsilon\sim\mathcal{N}(0,I).
\end{aligned}
\label{eq:haim_corruption}
\end{equation}
where $p_s$ controls feature masking and $\eta_s$ controls additive
Gaussian noise.

\begin{table}[t]
\centering
\caption{HAIM representation-level corruption parameters.}
\label{tab:haim_corruption}
\small
\begin{adjustbox}{max width=\linewidth}
\begin{tabular}{lcc}
\toprule
\textbf{Severity} & \textbf{Mask prob. $p_s$}
& \textbf{Noise SD $\eta_s$}\\
\midrule
Mild     & 0.10 & 0.10\\
Moderate & 0.30 & 0.30\\
Severe   & 0.50 & 0.50\\
\bottomrule
\end{tabular}
\end{adjustbox}
\end{table}

Exactly one modality is corrupted per condition. Four modalities and
three severity levels therefore give $4\times3=12$ corruption
conditions per model and seed.

\subsubsection{Matched Corruption Across Methods}
\label{app:matched_corruption}

For HAIM, corruption realizations are matched across competing methods.
The random corruption seed depends only on the experimental seed,
corrupted modality, and severity, and not on method identity. Thus, for
a given condition, all methods receive the same feature mask and
Gaussian-noise realization. This reduces variation caused by different
random perturbations and enables more direct robustness comparisons.

\subsection{Dataset and Cohort Details}
\label{app:data}

This section summarizes cohort construction, partitioning, and
preprocessing for the two evaluation datasets.

\subsubsection{Symile-MIMIC}
\label{app:symile_data}

We use Symile-MIMIC for cardiomegaly classification from three
modalities: chest radiographs (CXR), electrocardiograms (ECG), and
laboratory measurements. Per observation, the input shapes are
CXR ($3\times320\times320$), ECG ($1\times5000\times12$),
and laboratory measurements ($50$ features).
The released train, validation, and test partitions contain 10,000,
750, and 4,640 observations, respectively.

\paragraph{Label alignment and cohort construction.}
Not every released observation has a usable binary cardiomegaly label.
We construct a valid-label mask from the metadata and apply the same
mask to all modalities, laboratory missingness indicators, and
outcomes. Alignment is performed using the admission identifier
(\texttt{hadm\_id}) to preserve correspondence across modalities.
The resulting cohort is summarized in
Table~\ref{tab:symile_cohort}.

\begin{table}[t]
\centering
\caption{Symile-MIMIC cohort after label filtering.}
\label{tab:symile_cohort}
\small
\setlength{\tabcolsep}{4pt}
\begin{adjustbox}{max width=\linewidth}
\begin{tabular}{lrrr}
\toprule
\textbf{Split} & \textbf{Released} & \textbf{Usable}
& \textbf{Positive}\\
\midrule
Train      & 10,000 & 4,232 & 3,346\\
Validation & 750    & 349   & --\\
Test       & 4,640  & 1,918 & 1,444\\
\bottomrule
\end{tabular}
\end{adjustbox}
\end{table}

We retain the released train/validation/test partitioning. Admission
identifiers were checked across partitions before final evaluation,
with no overlap found between train, validation, and test sets.

\paragraph{Laboratory representation.}
Each observation contains 50 normalized laboratory values and 50
corresponding missingness indicators. These are concatenated into the
100-dimensional laboratory input described in
Section~\ref{app:model_architectures}. Missingness indicators are
retained when laboratory values are corrupted.

\subsubsection{HAIM-MIMIC-MM}
\label{app:haim_data}

We construct a HAIM-MIMIC-MM cohort for 48-hour mortality prediction
using four modality groups: structured data, physiological time-series,
CXR, and text. Their representation dimensions are 1,542, 451, 2,084,
and 2,304, respectively.

The cohort contains 45,036 multimodal observations from 6,486 unique
subjects, with 1,229 positive mortality observations
($2.73\%$ observation-level prevalence).

\paragraph{Patient-level partitioning.}
Because subjects may contribute multiple observations, splitting at
the observation level could introduce patient overlap across
partitions. We therefore construct train, validation, and test sets at
the subject level, assigning all observations from a subject to the
same partition.

\begin{table}[t]
\centering
\caption{HAIM 48-hour mortality cohort.}
\label{tab:haim_cohort}
\small
\setlength{\tabcolsep}{4pt}
\begin{adjustbox}{max width=\linewidth}
\begin{tabular}{lrrr}
\toprule
\textbf{Split} & \textbf{Obs.} & \textbf{Subjects}
& \textbf{Positive}\\
\midrule
Train      & 31,213 & 4,540 & 865\\
Validation & 5,059  & 648   & 118\\
Test       & 8,764  & 1,298 & 246\\
\bottomrule
\end{tabular}
\end{adjustbox}
\end{table}

There is no subject overlap between train, validation, and test
partitions, and no CXR image identifier is shared across partitions.

\paragraph{Preprocessing.}
HAIM experiments use distributed precomputed representations associated with CXR-anchored observations, with mortality defined over the subsequent 48 hours. Several physiological and imaging components are bounded to the anchor, but not every precomputed feature component can be guaranteed to share the same temporal boundary; we therefore interpret HAIM as a benchmark evaluation rather than a prospective deployment study. Normalization statistics are estimated on training data and applied unchanged to validation and test data.

\subsubsection{Comparison of Evaluation Settings}
\label{app:dataset_comparison}

The two datasets provide complementary evaluation settings. Symile
uses three modalities for an imaging-related finding, whereas HAIM
uses four heterogeneous modality groups for a patient outcome.
Consequently, Symile has $2^3-1=7$ non-empty modality configurations
(6 incomplete), while HAIM has $2^4-1=15$ configurations
(14 incomplete).

\begin{table}[t]
\centering
\caption{Summary of the evaluation settings.}
\label{tab:dataset_comparison}
\small
\setlength{\tabcolsep}{3pt}
\begin{tabularx}{\linewidth}{lXX}
\toprule
 & \textbf{Symile} & \textbf{HAIM}\\
\midrule
Task &
Cardiomegaly &
48-h mortality\\
Modalities &
CXR, ECG, labs &
Structured, TS, CXR, text\\
No. modalities &
3 & 4\\
Incomplete sets &
6 & 14\\
Split &
Released &
Patient-level\\
Input &
Raw/low-level &
Representations\\
\bottomrule
\end{tabularx}
\end{table}

\subsection{Dataset-Specific Implementation of Comparison Methods}
\label{app:baselines}

We compare with DrFuse, M$^3$Care, MUSE, and Flex-MoE in addition to
unimodal and conventional fusion baselines. The published methods assume
different modality configurations, input representations, and task-specific
encoders, so each is instantiated for Symile and HAIM using dataset-specific
input components while preserving its defining fusion and missing-modality
mechanism. Training follows the standard methodology of each approach rather
than imposing the proposed model's auxiliary objectives on competitors. All
methods use the same cohort partitions, outcome definitions, and held-out
evaluation conditions.

\subsubsection{DrFuse}
\label{app:drfuse}

DrFuse separates shared and modality-specific representations and uses
attention-based fusion with auxiliary alignment and disentanglement
objectives \citep{yao2024drfuse}. For Symile, we retain CXR as the
imaging branch and combine ECG and laboratory information into the
clinical branch, preserving shared/private projections, shared-space
alignment, disentanglement, attention, and auxiliary prediction losses.
For HAIM, the same shared/private principle is extended to structured,
time-series, CXR, and text representations, with unavailable modalities
masked from fusion. Dataset-specific encoders are used in both settings while retaining the defining DrFuse objectives and fusion mechanism.

\subsubsection{M$^3$Care}
\label{app:m3care}

M$^3$Care compensates for missing modalities using task-guided latent
representations retrieved from similar training examples
\citep{zhang2022m3care}. We retain this latent-memory mechanism for both
datasets, using Symile-specific modality encoders and
representation-level HAIM encoders. Missing representations are
estimated from neighboring examples using the modalities that remain
observed. For HAIM, the training-only memory contains at most 2,048
examples; the eight most similar entries are retrieved with temperature
$0.10$. Validation and test examples may query the frozen training
memory but are never inserted into it, preventing cross-partition
information leakage.

\subsubsection{MUSE}
\label{app:muse}

MUSE models available modalities as graph nodes and combines them
through learned message passing with contrastive graph-view
regularization \citep{wu2024multimodal}. We construct patient--modality
graphs from the three Symile or four HAIM modalities after projecting
dataset-specific inputs to a common latent space. Missing modalities
are excluded from the graph rather than represented by zero-valued
inputs. For HAIM, stochastic edge dropout generates contrastive graph
views, using edge-dropout probability $0.20$, contrastive weight $0.10$,
and temperature $0.20$. Thus, the implementation retains graph-based fusion, message passing, and graph-view consistency while using dataset-specific input encoders.

\subsubsection{Flex-MoE}
\label{app:flexmoe}

Flex-MoE handles arbitrary modality combinations through
missingness-aware sparse expert routing \citep{yun2024flexmoe}. The
Symile implementation retains the missing-modality bank,
combination-aware routing, and load-balancing objective, using 16
experts with top-$4$ routing and local CXR, ECG, and laboratory
encoders. Training begins with a five-epoch complete-modality warm-up.
For HAIM, the four modality representations are projected to a common
space and combined with an explicit modality-availability vector before
routing; this implementation uses eight experts with top-$2$ routing
and load-balancing weight $0.01$. Both settings preserve the central missingness-aware routing mechanism.

\subsubsection{Evaluation Fairness}
\label{app:baseline_fairness}

All comparison methods are independently trained over the same five experimental seeds and evaluated on identical held-out partitions and missing-modality conditions. Each method follows its standard methodological training mechanism: the proposed model uses its specified missing-subset schedule; modality dropout, M$^3$Care, and Flex-MoE retain their respective missingness augmentation or compensation procedures; conventional fusion baselines use complete-input training; and MUSE retains graph-view edge dropout. Our representation-level HAIM DrFuse implementation preserves its shared/private representation and fusion mechanism but handles unavailable modalities architecturally at evaluation rather than reproducing the original CXR-removal augmentation. We therefore interpret the benchmark as a comparison of complete learning methods under common data and evaluation conditions, while Table~\ref{tab:ablation} provides controlled within-framework component comparisons. No model in the final comparison is trained on the synthetic corruption operators used for corruption evaluation. For HAIM, methods additionally receive matched corruption realizations as described in Section~\ref{app:matched_corruption}. No held-out test examples are used for model selection or construction of training-time auxiliary structures.

\section{Results}
\label{sec:results}

We evaluate complete inputs, modality absence, and corruption, followed by validation-set component ablations and a post-hoc analysis of reliability and latent uncertainty. Complete condition-wise results are provided in the detailed results below.

\subsection{Complete-Modality Performance}
\label{sec:clean_results}

Table~\ref{tab:main_results} summarizes held-out AUROC under complete,
missing, and corrupted inputs. With complete modalities, the proposed model is comparable to the strongest methods in AUROC on both datasets ($0.7583$ on Symile; $0.8694$ on HAIM). On Symile it also has the highest mean AUPRC ($0.8856\pm0.0069$). On the highly imbalanced HAIM task, however, Flex-MoE obtains substantially higher AUPRC ($0.2943\pm0.0127$ versus $0.2427\pm0.0121$), so the proposed method does not dominate positive-class retrieval.

Thus, complete-input discrimination is competitive, while the more informative comparison for this framework is performance under incomplete evidence.

\begin{table}[t]
\centering
\caption{Held-out AUROC. Missing and Corrupt. average all incomplete configurations and modality--severity conditions, respectively; full results are in the main text.}
\label{tab:main_results}
\small
\setlength{\tabcolsep}{5pt}
\begin{adjustbox}{max width=\linewidth}
\begin{tabular}{lccc|ccc}
\toprule
& \multicolumn{3}{c|}{\textbf{Symile-MIMIC}}
& \multicolumn{3}{c}{\textbf{HAIM-MIMIC-MM}}\\
\cmidrule(lr){2-4}\cmidrule(lr){5-7}
\textbf{Method}
& \textbf{Clean} & \textbf{Missing} & \textbf{Corrupt.}
& \textbf{Clean} & \textbf{Missing} & \textbf{Corrupt.}\\
\midrule
Early Fusion
& 0.7290 & 0.6711 & 0.7277
& 0.8595 & 0.7536 & 0.8465\\

Gaussian PoE
& 0.7207 & 0.6679 & 0.7220
& 0.8498 & 0.7490 & 0.8369\\

Modality Dropout
& 0.7175 & 0.6670 & 0.7182
& 0.8622 & 0.7747 & 0.8497\\

DrFuse
& 0.7527 & 0.7009 & 0.7464
& 0.8542 & 0.7530 & 0.8429\\

Flex-MoE
& 0.7569 & 0.6934 & 0.7501
& 0.8687 & 0.7825 & \textbf{0.8589}\\

\textbf{Ours}
& \textbf{0.7583} & \textbf{0.7014} & \textbf{0.7590}
& \textbf{0.8694} & \textbf{0.7980} & 0.8558\\
\bottomrule
\end{tabular}
\end{adjustbox}
\end{table}

\subsection{Robustness to Missing Modalities}
\label{sec:missing_results}

Under missing modalities, Symile differences among the strongest methods are small: the proposed model has mean AUROC $0.7014$, compared with $0.7009$ for DrFuse and $0.6934$ for Flex-MoE. The advantage is not uniform across every configuration:
DrFuse performs better when CXR alone is missing, whereas Flex-MoE is
competitive when laboratory information is absent. The proposed model
is particularly strong when ECG or multiple modalities are unavailable,
including AUROCs of $0.7278\pm0.0046$ with ECG missing and
$0.7277\pm0.0141$ when both ECG and laboratory measurements are
missing.

The pattern is stronger on HAIM. Across all 14 incomplete modality
configurations, the proposed model achieves
$0.7980\pm0.0077$ mean AUROC, compared with
$0.7825\pm0.0096$ for Flex-MoE and
$0.7747\pm0.0129$ for modality dropout. Flex-MoE nevertheless obtains
the higher mean AUPRC ($0.1699\pm0.0061$ versus
$0.1614\pm0.0084$), showing that the advantage depends on the
evaluation metric. A paired seed-level analysis of the AUROC comparison with
Flex-MoE is provided in Section~\ref{app:haim_paired_missing}.

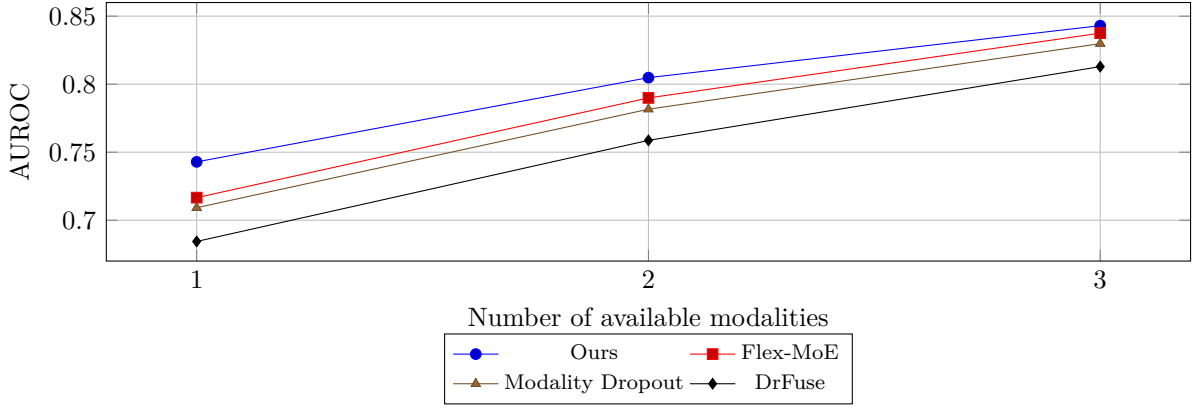
\begin{figure}[t]
\centering
\begin{tikzpicture}
\begin{axis}[
    width=\linewidth,
    height=5.0cm,
    xlabel={Number of available modalities},
    ylabel={AUROC},
    xmin=0.8, xmax=3.2,
    ymin=0.67, ymax=0.86,
    xtick={1,2,3},
    grid=major,
    legend style={
        font=\scriptsize,
        at={(0.5,-0.28)},
        anchor=north,
        legend columns=2
    },
    tick label style={font=\small},
    label style={font=\small},
]

\addplot+[mark=*]
coordinates {
    (1,0.7429)
    (2,0.8048)
    (3,0.8430)
};
\addlegendentry{Ours}

\addplot+[mark=square*]
coordinates {
    (1,0.7166)
    (2,0.7899)
    (3,0.8375)
};
\addlegendentry{Flex-MoE}

\addplot+[mark=triangle*]
coordinates {
    (1,0.7092)
    (2,0.7816)
    (3,0.8298)
};
\addlegendentry{Modality Dropout}

\addplot+[mark=diamond*]
coordinates {
    (1,0.6844)
    (2,0.7587)
    (3,0.8129)
};
\addlegendentry{DrFuse}

\end{axis}
\end{tikzpicture}
\caption{HAIM missing-modality robustness. AUROC averaged by number of available modalities within each seed.}
\label{fig:missing_robustness}
\end{figure}

Figure~\ref{fig:missing_robustness} shows how performance changes as
less evidence remains available on HAIM. The separation is most
pronounced in the more difficult incomplete settings: with only one
modality available, the proposed model reaches
$0.7429\pm0.0111$ AUROC, compared with
$0.7166\pm0.0064$ for Flex-MoE and
$0.7092\pm0.0105$ for modality dropout. With two and three modalities
available, it reaches $0.8048\pm0.0079$ and
$0.8430\pm0.0062$, respectively.

\begin{figure}[t]
    \centering
    \includegraphics[width=0.95\linewidth]
    {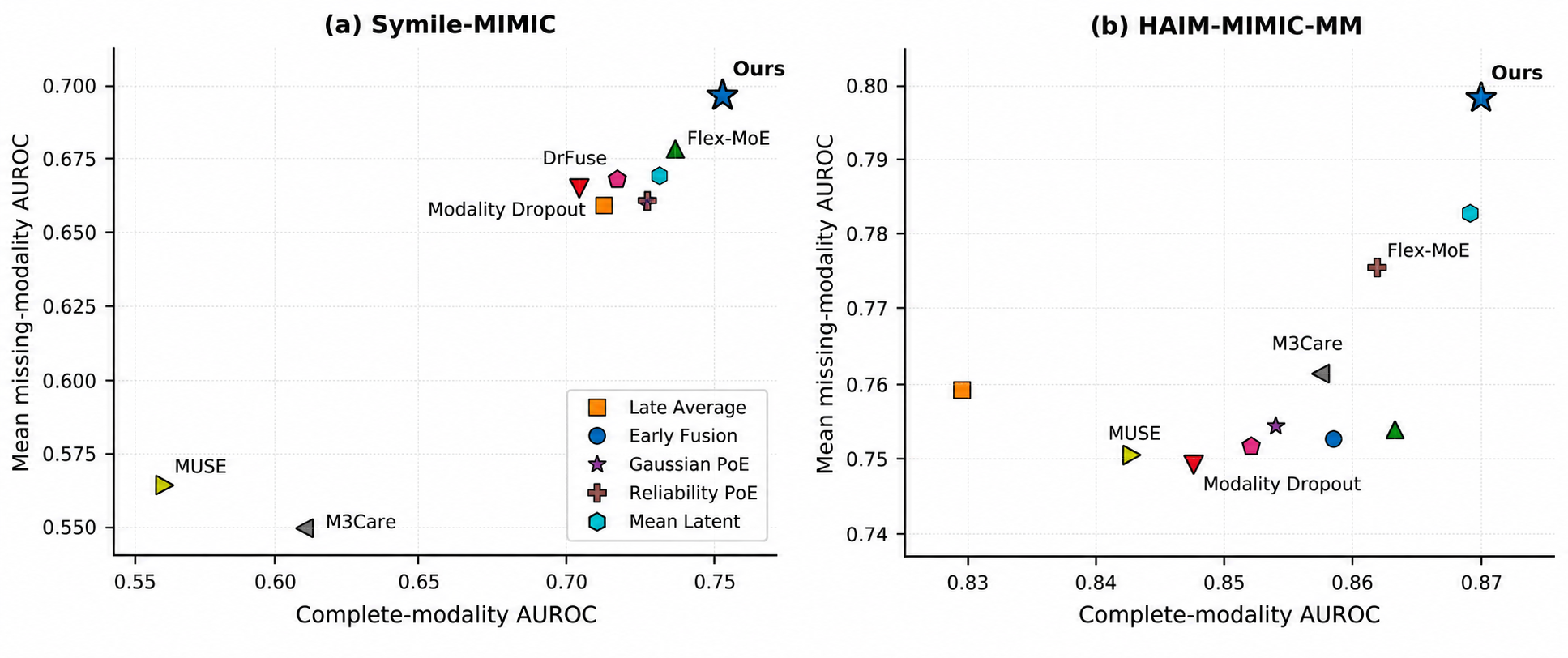}
    \caption{Clean versus missing-modality AUROC. Missing AUROC averages six Symile and 14 HAIM incomplete configurations.}
    \label{fig:clean_missing_tradeoff}
\end{figure}

Figure~\ref{fig:clean_missing_tradeoff} provides a complementary view
of this trade-off. On both datasets, the proposed model lies near the
upper-right of the clean-versus-missing performance plane. The
separation is particularly clear on HAIM: the proposed model and
Flex-MoE achieve nearly identical complete-modality AUROC, while the
proposed model retains higher mean AUROC when modalities are missing.

\subsection{Robustness to Corrupted Evidence}
\label{sec:corruption_results}

Corruption provides a diagnostic test of learned evidence weighting rather than a setting in which we claim uniform superiority.
On Symile, the proposed model remains highly stable across the nine
corruption conditions, with a mean AUROC of approximately $0.759$.
CXR corruption has little effect, while ECG corruption produces the
clearest degradation at severe intensity
($0.7447\pm0.0136$). Laboratory corruption does not lead to monotonic
AUROC degradation, emphasizing that increasing input-space corruption
does not necessarily imply monotonically increasing downstream task
difficulty.

On HAIM, the proposed model obtains a mean corrupted-input AUROC of
$0.8558\pm0.0066$, corresponding to an average decrease of
$0.0136$ from clean performance. Flex-MoE is strongest overall under
corruption ($0.8589\pm0.0115$) and exhibits a smaller mean decrease
($0.0097$). The proposed model degrades from
$0.8674\pm0.0058$ under mild corruption to
$0.8422\pm0.0100$ under severe corruption. Corruption sensitivity is
also strongly modality dependent: degradation of structured and
time-series information is substantially more harmful than comparable
perturbation of CXR or text representations.

Taken together, missingness and corruption are distinct problems. A missing expert can be removed exactly, whereas a corrupted expert remains present and must first be recognized as unreliable. The guarantees in Section~\ref{sec:theory} are therefore conditional on the learned reliability values and do not imply that every corruption will be detected.

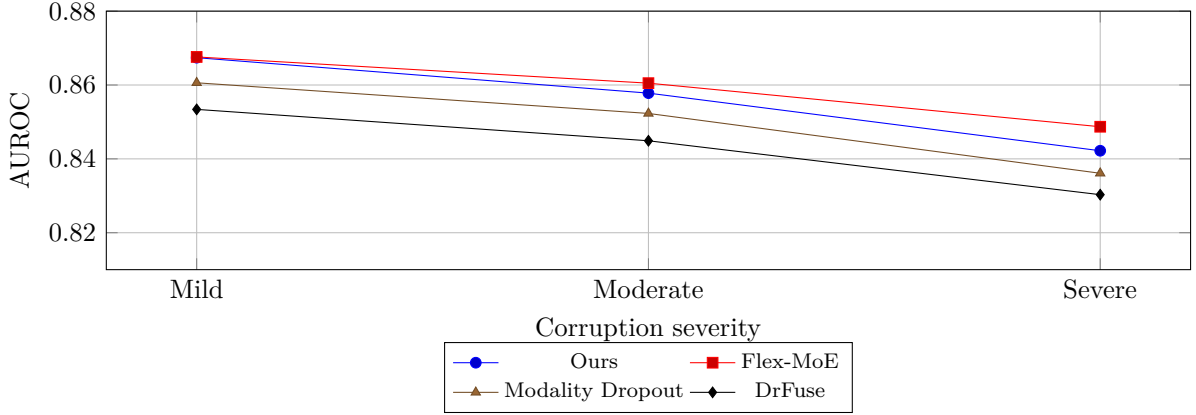
\begin{figure}[t]
\centering
\begin{tikzpicture}
\begin{axis}[
    width=\linewidth,
    height=5.0cm,
    xlabel={Corruption severity},
    ylabel={AUROC},
    symbolic x coords={Mild,Moderate,Severe},
    xtick=data,
    ymin=0.81, ymax=0.88,
    grid=major,
    legend style={
        font=\scriptsize,
        at={(0.5,-0.28)},
        anchor=north,
        legend columns=2
    },
    tick label style={font=\small},
    label style={font=\small},
]

\addplot+[mark=*]
coordinates {
    (Mild,0.8674)
    (Moderate,0.8578)
    (Severe,0.8422)
};
\addlegendentry{Ours}

\addplot+[mark=square*]
coordinates {
    (Mild,0.8676)
    (Moderate,0.8605)
    (Severe,0.8487)
};
\addlegendentry{Flex-MoE}

\addplot+[mark=triangle*]
coordinates {
    (Mild,0.8606)
    (Moderate,0.8523)
    (Severe,0.8361)
};
\addlegendentry{Modality Dropout}

\addplot+[mark=diamond*]
coordinates {
    (Mild,0.8534)
    (Moderate,0.8449)
    (Severe,0.8303)
};
\addlegendentry{DrFuse}

\end{axis}
\end{tikzpicture}
\caption{HAIM corruption robustness. AUROC averaged over corrupted modalities within each seed.}
\label{fig:corruption_robustness}
\end{figure}

\subsection{Reliability and Uncertainty under Corruption}
\label{sec:reliability_analysis}

\begin{table}[t]
\centering
\caption{Validation ablation. Missing is mean AUROC over incomplete configurations.}
\label{tab:ablation}
\small
\setlength{\tabcolsep}{4pt}
\begin{adjustbox}{max width=\linewidth}
\begin{tabular}{lccccc|cc|cc}
\toprule
& \multicolumn{5}{c|}{\textbf{Components}}
& \multicolumn{2}{c|}{\textbf{Symile}}
& \multicolumn{2}{c}{\textbf{HAIM}}\\
\cmidrule(lr){2-6}
\cmidrule(lr){7-8}
\cmidrule(lr){9-10}
\textbf{Model}
& \textbf{Mean}
& \textbf{PoE}
& \textbf{Rel.}
& \textbf{Miss.}
& \textbf{Cons.}
& \textbf{Clean}
& \textbf{Missing}
& \textbf{Clean}
& \textbf{Missing}\\
\midrule
A & \checkmark & & & & &
0.7750 & 0.7210 &
0.8960 & 0.7791\\

B & & \checkmark & & & &
0.7766 & 0.7122 &
0.8983 & 0.7856\\

C & & \checkmark & \checkmark & & &
0.7797 & 0.7187 &
0.9013 & 0.7876\\

D & & \checkmark & & \checkmark & &
0.7818 & 0.7198 &
0.9023 & 0.8120\\

E & & \checkmark & \checkmark & \checkmark & &
0.7721 & 0.7127 &
0.9042 & 0.8095\\

D+C & & \checkmark & & \checkmark & \checkmark &
0.7830 & 0.7199 &
0.9062 & 0.8104\\

F & & \checkmark & \checkmark & \checkmark & \checkmark &
\textbf{0.7835} & \textbf{0.7210} &
\textbf{0.9090} & \textbf{0.8251}\\
\bottomrule
\end{tabular}
\end{adjustbox}
\end{table}

We next inspect frozen checkpoints without retraining or model selection. For each corrupted modality we summarize the scalar reliability $r_m$, mean latent variance $d^{-1}\sum_j\sigma_{mj}^2$, and mean effective precision $d^{-1}\sum_j r_m/\sigma_{mj}^2$. On Symile, laboratory corruption gives the clearest change in the learned reliability weight: from clean to severe corruption, mean reliability falls from $0.366$ to $0.107$, latent variance rises from $2.98$ to $5.85$, and effective precision falls from $0.140$ to $0.027$. ECG reliability decreases through moderate corruption ($0.447$ to $0.357$) but rises under severe corruption ($0.513$) despite lower AUROC, exposing a failure mode under strong degradation. CXR reliability changes little and slightly increases, consistent with the negligible downstream effect of the applied CXR corruption.

HAIM shows a similarly modality-dependent pattern. Severe structured corruption increases latent variance from $1.11$ to $1.36$ and decreases reliability from $0.951$ to $0.944$, while time-series and text variance increase but reliability remains stable or rises slightly. Because these corruptions are unseen during training, the analysis tests whether task-learned evidence weights respond to degradation without explicit corruption supervision. They do so inconsistently: reliability weights and latent uncertainty exhibit distinct responses, and $r_m$ should not be interpreted as a universally calibrated corruption score. Full values are reported in Section~\ref{app:reliability_analysis}.

\subsection{Component Ablation}
\label{sec:ablation}

Table~\ref{tab:ablation} reports component ablations on
\emph{validation data}, separately from the final held-out test
evaluation above. The ablations are averaged over five independently
trained seeds. Accordingly, these results are intended to identify component behavior rather than provide additional held-out comparisons. D+C is a post-hoc development control and was not evaluated on the held-out test set.

On Symile, component effects are modest: missing-aware training improves Gaussian PoE under modality absence, while the full model attains the highest clean and mean-missing validation AUROC ($0.7835$ and $0.7210$).

The HAIM development ablation more clearly separates the contributions. Missing-aware training increases mean incomplete AUROC from $0.7856$ for Gaussian PoE (B) to $0.8120$ (D). Adding consistency without reliability (D+C) gives $0.8104$, whereas adding reliability under otherwise matched PoE, missing-aware, and consistency training (D+C $\rightarrow$ F) increases mean incomplete AUROC to $0.8251$ and clean AUROC from $0.9062$ to $0.9090$. Thus, missing-aware training provides a substantial robustness gain, and reliability weighting provides an additional benefit in the controlled HAIM comparison.

\subsection{Additional Symile-MIMIC Results}
\label{app:symile_results}

This section provides the detailed held-out Symile-MIMIC results
summarized in Section~\ref{sec:results}. Unless stated otherwise,
results are mean $\pm$ sample standard deviation over five independently
trained seeds. Higher AUROC and AUPRC are better; lower Brier score is
better.

\subsubsection{Complete-Modality Performance}
\label{app:symile_clean}

\begin{table}[t]
\centering
\caption{Symile-MIMIC held-out performance with all modalities
available. Results are mean $\pm$ SD over five seeds.}
\label{tab:symile_clean_full}
\small
\setlength{\tabcolsep}{5pt}
\begin{adjustbox}{max width=\linewidth}
\begin{tabular}{lccc}
\toprule
\textbf{Method} & \textbf{AUROC} & \textbf{AUPRC} & \textbf{Brier}\\
\midrule
CXR-only
& $0.7133\pm0.0123$ & $0.8612\pm0.0056$ & $0.2474\pm0.1159$\\
ECG-only
& $0.6452\pm0.0236$ & $0.8280\pm0.0172$ & $0.2219\pm0.0189$\\
Labs-only
& $0.6201\pm0.0050$ & $0.8271\pm0.0036$ & $0.2546\pm0.0435$\\
Early Fusion
& $0.7290\pm0.0193$ & $0.8585\pm0.0077$ & $0.2312\pm0.0759$\\
Late Average
& $0.7046\pm0.0303$ & $0.8569\pm0.0160$ & $0.2321\pm0.0371$\\
Mean Latent
& $0.7370\pm0.0135$ & $0.8713\pm0.0074$ & $0.2536\pm0.0967$\\
Gaussian PoE
& $0.7207\pm0.0154$ & $0.8638\pm0.0134$ & $0.2299\pm0.0773$\\
Reliability PoE
& $0.7246\pm0.0173$ & $0.8596\pm0.0064$ & $0.2096\pm0.0326$\\
Modality Dropout
& $0.7175\pm0.0228$ & $0.8551\pm0.0110$ & $0.2276\pm0.0400$\\
DrFuse
& $0.7527\pm0.0135$ & $0.8770\pm0.0066$ & $\mathbf{0.1529\pm0.0086}$\\
M$^3$Care
& $0.6097\pm0.0181$ & $0.8093\pm0.0130$ & $0.2500\pm0.0184$\\
MUSE
& $0.5609\pm0.0709$ & $0.7955\pm0.0274$ & $0.1844\pm0.0044$\\
Flex-MoE
& $0.7569\pm0.0214$ & $0.8709\pm0.0114$ & $0.2168\pm0.0352$\\
\textbf{Ours}
& $\mathbf{0.7583\pm0.0106}$
& $\mathbf{0.8856\pm0.0069}$
& $0.2184\pm0.0349$\\
\bottomrule
\end{tabular}
\end{adjustbox}
\end{table}

\subsubsection{Missing-Modality Performance}
\label{app:symile_missing}

Table~\ref{tab:symile_missing_full} reports AUROC for every non-empty incomplete modality configuration. Unimodal entries reproduce the corresponding held-out unimodal evaluation from Table~\ref{tab:symile_clean_full}; N/A denotes absence of the required modality.

\begin{table}[t]
\centering
\caption{Symile-MIMIC held-out AUROC under missing modalities.
Results are mean $\pm$ SD over five seeds.}
\label{tab:symile_missing_full}
\scriptsize
\setlength{\tabcolsep}{2.8pt}
\begin{adjustbox}{max width=\linewidth}
\begin{tabular}{lcccccc}
\toprule
\textbf{Method}
& \textbf{CXR miss.}
& \textbf{ECG miss.}
& \textbf{Labs miss.}
& \textbf{CXR+ECG}
& \textbf{CXR+Labs}
& \textbf{ECG+Labs}\\
\midrule
CXR-only
& N/A & $0.7133\pm.0123$ & $0.7133\pm.0123$
& N/A & N/A & $0.7133\pm.0123$\\
ECG-only
& $0.6452\pm.0236$ & N/A & $0.6452\pm.0236$
& N/A & $0.6452\pm.0236$ & N/A\\
Labs-only
& $0.6201\pm.0050$ & $0.6201\pm.0050$ & N/A
& $0.6201\pm.0050$ & N/A & N/A\\
Early Fusion
& $0.6512\pm.0256$ & $0.7031\pm.0090$ & $0.7295\pm.0235$
& $0.6135\pm.0089$ & $0.6290\pm.0175$ & $0.7005\pm.0167$\\
Late Average
& $0.6556\pm.0229$ & $0.6914\pm.0335$ & $0.7057\pm.0393$
& $0.6217\pm.0139$ & $0.6263\pm.0385$ & $0.6996\pm.0140$\\
Mean Latent
& $0.6532\pm.0220$ & $0.7077\pm.0102$ & $0.7352\pm.0053$
& $0.6114\pm.0130$ & $0.6287\pm.0303$ & $0.7087\pm.0107$\\
Gaussian PoE
& $0.6472\pm.0136$ & $0.6953\pm.0075$ & $0.7286\pm.0097$
& $0.6089\pm.0070$ & $0.6248\pm.0163$ & $0.7026\pm.0102$\\
Reliability PoE
& $0.6510\pm.0251$ & $0.6978\pm.0100$ & $0.7281\pm.0154$
& $0.6146\pm.0107$ & $0.6181\pm.0419$ & $0.7026\pm.0095$\\
Modality Dropout
& $0.6488\pm.0317$ & $0.6974\pm.0098$ & $0.7271\pm.0182$
& $0.6157\pm.0095$ & $0.6130\pm.0402$ & $0.7001\pm.0072$\\
DrFuse
& $\mathbf{0.7147\pm.0192}$ & $0.7036\pm.0091$ & $0.7590\pm.0118$
& $0.6179\pm.0057$ & $\mathbf{0.7023\pm.0179}$ & $0.7078\pm.0048$\\
M$^3$Care
& $0.6097\pm.0181$ & $0.6122\pm.0186$ & $0.4733\pm.0438$
& $0.6122\pm.0186$ & $0.4839\pm.0503$ & $0.5013\pm.0773$\\
MUSE
& $0.4796\pm.0260$ & $0.6491\pm.0539$ & $0.5907\pm.0735$
& $0.5688\pm.0494$ & $0.4820\pm.0360$ & $0.6332\pm.0580$\\
Flex-MoE
& $0.6877\pm.0318$ & $0.7155\pm.0103$ & $0.7647\pm.0177$
& $0.6118\pm.0218$ & $0.6733\pm.0399$ & $0.7071\pm.0137$\\
\textbf{Ours}
& $0.6853\pm.0200$
& $\mathbf{0.7278\pm.0046}$
& $\mathbf{0.7655\pm.0123}$
& $0.6127\pm.0190$
& $0.6891\pm.0163$
& $\mathbf{0.7277\pm.0141}$\\
\bottomrule
\end{tabular}
\end{adjustbox}
\end{table}

The proposed model does not dominate every configuration: DrFuse is strongest when CXR or CXR+Labs are missing, while the proposed model has mean AUROC $0.7014$ across the six incomplete conditions and is strongest when ECG, Labs, or ECG+Labs are absent.

\subsubsection{Corruption Robustness}
\label{app:symile_corruption_results}

For compactness, Table~\ref{tab:symile_corruption_full} reports AUROC
under each modality and severity combination. Complete AUPRC and Brier
values follow the same evaluation protocol.

\begin{table}[t]
\centering
\caption{Symile-MIMIC held-out AUROC under corruption.
$s_1$, $s_2$, and $s_3$ denote mild, moderate, and severe corruption.}
\label{tab:symile_corruption_full}
\scriptsize
\setlength{\tabcolsep}{2.8pt}
\begin{adjustbox}{max width=\linewidth}
\begin{tabular}{lccc|ccc|ccc}
\toprule
& \multicolumn{3}{c|}{\textbf{CXR}}
& \multicolumn{3}{c|}{\textbf{ECG}}
& \multicolumn{3}{c}{\textbf{Labs}}\\
\textbf{Method}
& $s_1$ & $s_2$ & $s_3$
& $s_1$ & $s_2$ & $s_3$
& $s_1$ & $s_2$ & $s_3$\\
\midrule
Early Fusion
& .7293 & .7298 & .7306 & .7267 & .7220 & .7148
& .7317 & .7331 & .7316\\
Late Average
& .7046 & .7044 & .7080 & .7032 & .6972 & .6900
& .7093 & .7144 & .7062\\
Mean Latent
& .7368 & .7367 & .7376 & .7365 & .7358 & .7305
& .7395 & .7413 & .7361\\
Gaussian PoE
& .7210 & .7220 & .7224 & .7200 & .7175 & .7119
& .7248 & .7302 & .7282\\
Reliability PoE
& .7239 & .7242 & .7253 & .7225 & .7105 & .6950
& .7300 & .7327 & .7259\\
Modality Dropout
& .7177 & .7189 & .7210 & .7163 & .7121 & .7034
& .7220 & .7269 & .7251\\
DrFuse
& .7520 & .7497 & .7518 & .7511 & .7292 & .7123
& .7564 & .7599 & .7555\\
M$^3$Care
& .6097 & .6097 & .6097 & .6097 & .6096 & .6093
& .6109 & .5963 & .5493\\
MUSE
& .5462 & .5476 & .5480 & .5387 & .5413 & .5738
& .5467 & .5465 & .5464\\
Flex-MoE
& .7574 & .7590 & .7569 & .7513 & .7266 & .7112
& .7625 & .7653 & .7608\\
\textbf{Ours}
& .7584 & .7570 & .7574
& .7576 & .7553 & .7447
& .7645 & .7678 & .7687\\
\bottomrule
\end{tabular}
\end{adjustbox}
\end{table}

For the proposed model, CXR corruption produces little change from
clean performance, while severe ECG corruption gives the clearest
degradation. Laboratory corruption does not produce monotonic AUROC
degradation, illustrating that increasing synthetic input distortion
does not necessarily translate to increasing downstream prediction
difficulty.

\subsection{Reliability--Uncertainty Analysis}
\label{app:reliability_analysis}

This post-hoc analysis uses frozen checkpoints only; no additional training, validation evaluation, or model selection is performed. For each modality and corruption severity, we report the mean reliability $r_m$, mean latent variance $d^{-1}\sum_j\sigma_{mj}^2$, and mean effective precision $d^{-1}\sum_j r_m/\sigma_{mj}^2$. The analysis is descriptive: the fusion guarantees are conditional on $r_m$ and do not require monotonic corruption detection.

\begin{table}[t]
\centering
\caption{Symile reliability and uncertainty under corruption. Values are mean $\pm$ SD over test observations for the analyzed frozen checkpoint.}
\label{tab:symile_rel_unc}
\scriptsize
\setlength{\tabcolsep}{4pt}
\begin{adjustbox}{max width=\linewidth}
\begin{tabular}{llccc}
\toprule
Modality & Severity & Reliability $r_m$ & Mean variance & Effective precision \\\\
\midrule
CXR & Clean & $0.5538\pm0.1280$ & $1.0221\pm0.3697$ & $0.5658\pm0.1250$ \\\\
 & Mild & $0.5596\pm0.1258$ & $1.0265\pm0.3828$ & $0.5707\pm0.1229$ \\\\
 & Moderate & $0.5649\pm0.1223$ & $1.0300\pm0.3981$ & $0.5760\pm0.1188$ \\\\
 & Severe & $0.5702\pm0.1159$ & $1.0072\pm0.3578$ & $0.5898\pm0.1091$ \\\\
ECG & Clean & $0.4474\pm0.1861$ & $5.3194\pm3.8315$ & $0.1232\pm0.0921$ \\\\
 & Mild & $0.3963\pm0.1719$ & $5.2969\pm3.9904$ & $0.1119\pm0.0854$ \\\\
 & Moderate & $0.3566\pm0.1677$ & $4.6269\pm4.1334$ & $0.1079\pm0.0576$ \\\\
 & Severe & $0.5128\pm0.1344$ & $3.3373\pm3.5488$ & $0.2299\pm0.1044$ \\\\
Labs & Clean & $0.3663\pm0.1206$ & $2.9754\pm1.0118$ & $0.1398\pm0.0691$ \\\\
 & Mild & $0.3038\pm0.0969$ & $3.0978\pm1.0185$ & $0.1125\pm0.0591$ \\\\
 & Moderate & $0.1962\pm0.0819$ & $3.9593\pm1.5450$ & $0.0626\pm0.0443$ \\\\
 & Severe & $0.1066\pm0.0698$ & $5.8508\pm2.5667$ & $0.0271\pm0.0284$ \\\\
\bottomrule
\end{tabular}
\end{adjustbox}
\end{table}

For HAIM, the corresponding clean-to-severe changes are: structured reliability $0.9513\rightarrow0.9440$, variance $1.111\rightarrow1.359$, and effective precision $7.816\rightarrow7.405$; time-series reliability $0.9145\rightarrow0.9217$ and variance $1.538\rightarrow1.657$; CXR reliability $0.8872\rightarrow0.8878$ with nearly unchanged variance; and text reliability $0.8876\rightarrow0.8916$ with variance $1.344\rightarrow1.438$. These results reinforce that task-learned reliability is not a monotonic corruption score under unseen degradation.

\subsubsection{Paired Seed-Level Analysis of HAIM Missing-Modality AUROC}
\label{app:haim_paired_missing}

The main HAIM missing-modality result averages AUROC over the same 14 non-empty incomplete modality configurations within each experimental seed. Because the proposed model and Flex-MoE are evaluated over matched seeds and the same held-out conditions, we additionally summarize their seed-level paired differences. This analysis is supplementary to the mean $\pm$ sample SD reporting in the main text.

\begin{table}[!ht]
\centering
\caption{Paired seed-level HAIM missing-modality AUROC. Each value is the mean AUROC over the 14 incomplete modality configurations within one experimental seed. $\Delta$ denotes Ours minus Flex-MoE.}
\vspace{0.2cm}
\label{tab:haim_missing_paired}
\small
\setlength{\tabcolsep}{4pt}
\begin{adjustbox}{max width=\linewidth}
\begin{tabular}{rrrr}
\toprule
\textbf{Seed} & \textbf{Ours} & \textbf{Flex-MoE} & $\boldsymbol{\Delta}$\\\\
\midrule
0 & 0.8031 & 0.7880 & +0.0151\\\\
1 & 0.8008 & 0.7965 & +0.0043\\\\
2 & 0.8022 & 0.7764 & +0.0258\\\\
3 & 0.7845 & 0.7787 & +0.0058\\\\
4 & 0.7996 & 0.7731 & +0.0265\\\\
\midrule
Mean & 0.7980 & 0.7825 & +0.0155\\\\
\bottomrule
\end{tabular}
\end{adjustbox}
\end{table}

The paired AUROC difference is positive in all five seeds. The mean paired improvement is $+0.0155$ AUROC. Using a two-sided 95\% Student-$t$ confidence interval over the five seed-level paired differences gives an interval of approximately $[+0.0022,+0.0288]$. We report the interval as a descriptive uncertainty summary over independent training runs; the experimental seed, rather than the individual modality configuration, is treated as the unit of replication.

\subsection{Additional HAIM-MIMIC-MM Results}
\label{app:haim_results}

This section provides detailed held-out HAIM-MIMIC-MM results.
All values are aggregated over five independently trained seeds.
For missingness and corruption summaries, conditions are first
averaged within each seed before computing mean $\pm$ sample standard
deviation across seeds.

\subsubsection{Complete-Modality Performance}
\label{app:haim_clean}

\begin{table}[!htbp]
\centering
\caption{HAIM held-out performance with all four modalities
available}
\vspace{0.2cm}
\label{tab:haim_clean_full}
\small
\setlength{\tabcolsep}{5pt}
\begin{adjustbox}{max width=\linewidth}
\begin{tabular}{lccc}
\toprule
\textbf{Method} & \textbf{AUROC} & \textbf{AUPRC} & \textbf{Brier}\\
\midrule
Structured-only
& $0.7471\pm0.0160$ & $0.0801\pm0.0038$ & $0.1568\pm0.0339$\\
Time-series-only
& $0.7928\pm0.0073$ & $0.2120\pm0.0235$ & $0.1607\pm0.0202$\\
CXR-only
& $0.7080\pm0.0175$ & $0.0697\pm0.0105$ & $0.1790\pm0.0239$\\
Text-only
& $0.7175\pm0.0110$ & $0.0793\pm0.0056$ & $0.1502\pm0.0402$\\
Early Fusion
& $0.8595\pm0.0080$ & $0.2452\pm0.0309$ & $0.0672\pm0.0152$\\
Late Average
& $0.8300\pm0.0069$ & $0.1721\pm0.0209$ & $0.0974\pm0.0143$\\
Mean Latent
& $0.8637\pm0.0079$ & $0.2370\pm0.0211$ & $0.0663\pm0.0056$\\
Gaussian PoE
& $0.8498\pm0.0100$ & $0.2341\pm0.0201$ & $0.0737\pm0.0156$\\
Reliability PoE
& $0.8519\pm0.0117$ & $0.2267\pm0.0274$ & $0.0685\pm0.0110$\\
Modality Dropout
& $0.8622\pm0.0157$ & $0.2533\pm0.0169$ & $0.0687\pm0.0187$\\
DrFuse
& $0.8542\pm0.0145$ & $0.2479\pm0.0378$ & $0.0688\pm0.0153$\\
M$^3$Care
& $0.8584\pm0.0079$ & $0.2340\pm0.0137$ & $0.0820\pm0.0212$\\
MUSE
& $0.8455\pm0.0213$ & $0.2589\pm0.0161$ & $\mathbf{0.0596\pm0.0127}$\\
Flex-MoE
& $0.8687\pm0.0123$ & $\mathbf{0.2943\pm0.0127}$ & $0.0732\pm0.0091$\\
\textbf{Ours}
& $\mathbf{0.8694\pm0.0058}$ & $0.2427\pm0.0121$ & $0.0739\pm0.0505$\\
\bottomrule
\end{tabular}
\end{adjustbox}
\end{table}

\section{Discussion}
\label{sec:discussion}

The clearest gain is HAIM mean AUROC under missing evidence, with competitive complete-input performance. Missing-aware training provides the largest initial robustness gain, while reliability adds benefit in the matched five-seed HAIM development ablation. Under unseen corruption, learned weights respond inconsistently: Symile laboratory weights decrease strongly, severe ECG corruption produces a failure case, and HAIM weights are comparatively stable while Flex-MoE has higher mean corruption AUROC. Thus, the fusion rule provides explicit evidence control, but learning when to suppress degraded evidence remains unresolved.

\section{Limitations}
\label{sec:limitations}
The reliability weight and expert variance jointly determine effective precision, so task supervision does not identify $r_m$ as a calibrated quality score. Reliability-weight responses can fail under severe degradation. HAIM uses distributed precomputed representations; not every feature component is guaranteed to share the CXR anchor's temporal boundary. Evaluation is further limited to two clinical benchmarks, synthetic corruptions, dataset-specific implementations of comparison methods, and no output-level calibration analysis of fused latent variance.

\section{Conclusion}
\label{sec:conclusion}

We introduced reliability-aware Gaussian PoE fusion with explicit input-dependent evidence weighting and missing-expert omission. Across two clinical benchmarks, the framework is competitive with complete inputs and shows its clearest mean-AUROC advantage under modality absence. Missing-aware training drives much of this robustness, with an additional reliability benefit in the matched HAIM development ablation. Unseen corruption exposes both useful responses and failure cases, leaving dependable quality estimation open.

\bibliographystyle{unsrtnat}
\bibliography{refs}

\section*{AI Use Statement}
OpenAI's GPT-5.6 Sol was used to assist with manuscript editing,
clarity, organization, and \LaTeX{} preparation. All scientific
claims, methods, analyses, experimental results, and conclusions were
reviewed and verified by the authors, who take full responsibility for
the content of the paper.
\end{document}